\documentclass[11pt,a4paper]{article}
\usepackage[hyperref]{emnlp-ijcnlp-2019}
\usepackage{times}
\usepackage{latexsym}
\usepackage{graphicx}
\usepackage{multirow}
\usepackage{mwe}
\usepackage{xcolor}
\usepackage{microtype}
\usepackage{url}

\newcommand{\enotesoff}{\long\gdef\enote##1{}}
\newcommand{\enoteson}{\long\gdef\enote##1{\par\noindent\fbox{\parbox {0.48\textwidth}{{\large DRAFT.} \small\scshape ##1}}\\[0.3ex]}}
\newcommand{\enotesonabs}{\long\gdef\enoteabs##1{\par\noindent\fbox{\parbox {0.40\textwidth}{{\large DRAFT.} \small\scshape ##1}}\\[0.3ex]}}
\enotesoff
\enoteson
\enotesonabs

\aclfinalcopy % Uncomment this line for the final submission
\newcommand{\comment}[1]{}

\title{Controlled Text Generation for Data Augmentation in Intelligent Artificial Agents}

\author{Nikolaos Malandrakis$^1$, Minmin Shen$^2$, Anuj Goyal$^2$,\\ \textbf{Shuyang Gao$^2$, Abhishek Sethi$^2$, Angeliki Metallinou$^2$ }\\
  $^1$ Signal Analysis and Interpretation Laboratory (SAIL), USC, Los Angeles, CA 90089 \\
  $^2$ Amazon Alexa AI \\
  {\tt malandra@usc.edu,}\\ {\tt \{shenm,anujgoya,shuyag,abhsethi,ametalli\}@amazon.com}}

\date{}

\begin{document}
\maketitle

\begin{abstract}
%%%NOTE: not sure whether to adopt my abstraction,
Data availability is a bottleneck during early stages of development of new capabilities for intelligent artificial agents. We investigate the use of text generation techniques to augment the training data of a popular commercial artificial agent across categories of functionality, with the goal of faster development of new functionality.  We explore a variety of encoder-decoder generative models for synthetic training data generation and propose using  conditional variational auto-encoders. Our approach requires only direct optimization, works well with limited data and significantly outperforms the previous controlled text generation techniques.
%We also propose a novel evaluation framework that examines the generated data in terms of accuracy, diversity and novelty. 
Further, the generated data are used as additional training samples in an extrinsic intent classification task, leading to improved performance by up to 5\% absolute f-score in low-resource cases, validating the usefulness of our approach.
%These capabilities are typically defined via a small amount of phrase templates, which are later expanded via manual annotations. 

\end{abstract}

\section{Introduction}
\label{sec:introduction}
%\footnote{This work was performed while Nikolaos Malandrakis was interning at Amazon.}

Voice-powered artificial agents have seen widespread commercial use in recent years, with agents like Google's Assistant, Apple's Siri and Amazon's Alexa rising in popularity. These agents are  expected to be highly accurate in understanding the users' requests and to be capable of handling a variety of continuously expanding functionality.  New capabilities are initially defined via a few phrase templates. Those are expanded, typically through larger scale data collection, to create datasets for building the machine learning algorithms required to create a serviceable Natural Language Understanding (NLU) system. This is a lengthy and expensive process that is repeated for new functionality expansion and can significantly slow down development time.

%e.g., the phrase template `is \textbf{movie\_title} suitable for children' can be used as an example request for the a movie rating, in movie related functionality. 

We investigate the use of neural generative encoder-decoder models for text data generation.
Given a small set of phrase templates for some new functionality, our goal is to generate new \textit{semantically similar} phrases and augment our training data. This data augmentation is not necessarily meant as a replacement for large-scale data collection, but rather as a way to accelerate the early stages of new functionality development.
This task shares similarities with paraphrasing. Therefore, inspired by work in paraphrasing \cite{Prakash} and controlled text generation \cite{Hu2017}, we investigate the use of variational autoencoder models and methods to condition neural generators.% towards the target NLU functionality.

For controlled text generation, \cite{Hu2017} used a variational autoencoder with an additional discriminator and trained the model in a wake-sleep way. \cite{zhou2018} used reinforcement via an emoji classifier to generate emotional responses. However, we found that when the number of samples is relatively small compared to the number of categories, such an approach might be counter-productive, because the required classifier components can not perform well. Inspired by recent advantages of connecting information theory with variational auto-encoders and invariant feature learning \cite{Moyer2018}, we instead use this approach to our controlled text generation task, without a discriminator.

Furthermore, our task differs from typical paraphrasing in that semantic similarity between the output text and the NLU functionality is not the only objective. The synthetic data should be evaluated in terms of its lexical diversity and novelty, which are important properties of a high quality training set.

Our key contributions are as follows:

\begin{itemize}
\item We thoroughly investigate text generation techniques for NLU data augmentation with sequence to sequence model and variational auto-encoders, in an atypically low-resource setting.
%\item We introduce a new method for controlled text generation using conditional auto-encoder and invariant feature learning. We show that it significantly outperforms the previous method with wake-sleep training. 
%\item We propose a framework for evaluating intrinsic quality metrics of the generated data, such as accuracy, diversity and novelty.
\item We validate our method in an extrinsic intent classification task, showing that the generated data brings considerable accuracy gains in low resource settings. 
\end{itemize}

%To summarize our contributions, we describe a text data generation framework for NLU that can be used in the early stages of development and an evaluation framework more suitable to this generation task. We investigate multiple methods of text generation and evaluate on a large scale with multiple capabilities of a commercial virtual agent. We show that the method produces relevant data and leads to improved classification performance of an intent classifier, which should enable faster development of new models by reducing the reliance on large annotated datasets.

\section{Related Work}
\label{sec:related_work}

%[This section needs to be expanded]
%Our task is controlled, category-aware, text generation. It is a similar task to paraphrasing in that we want to generate new phrases similar to existing phrases, but unlike paraphrasing our problem is many inputs to many outputs, since all phrases within a category are considered equivalent.

Neural networks have revolutionized the field of text generation, in machine translation \cite{suts2014}, summarization \cite{see2017} and image captioning \cite{you2016}. However, conditional text generation has been relatively less studied as compared to conditional image generation and poses some unique problems. One of the issues is the non-differentiability of the sampled text that limits the applicability of a global discriminator in end-to-end training. The problem has been relatively addressed by using CNNs for generation \cite{cnn2017}, policy gradient reinforcement learning methods including SeqGAN \cite{yu2017}, LeakGAN \cite{guo2018}, or using latent representation like Gumbel softmax (\cite{gumbel2016}). Many of these approaches suffer from high training variance, mode collapse or cannot be evaluated beyond a qualitative analysis.

Many models have been proposed for text generation. Seq2seq models are standard encoder-decoder models widely used in text applications like machine translation \cite{Luong2015} and paraphrasing \cite{Prakash}. Variational Auto-Encoder (VAE) models are another important family \cite{Kingma2013} and they consist of an encoder that maps each sample to a latent representation and a decoder that generates samples from the latent space. The advantage of these models is the variational component and its potential to add diversity to the generated data. They have been shown to work well for text generation \cite{Bowman2016}. Conditional VAE (CVAE) \cite{kingma2014semi} was proposed to improve over seq2seq models for generating more diverse and relevant text. CVAE based models \cite{serban2017hierarchical, zhao2017learning, shen2017conditional,zhou2018} incorporate stochastic latent variables that represents the generated text, and append the output of VAE as an additional input to decoder.

Paraphrasing can be performed using neural networks with an encoder-decoder configuration, including sequence to sequence (S2S) \cite{Luong2015} and generative models \cite{Bowman2016} and various modifications have been proposed to allow for control of the output distribution of the data generation \cite{Yan2015,Hu2017}. 

Unlike the typical paraphrasing task we care about the lexical diversity and novelty of the generated output. This has been a concern in paraphrase generation: a generator that only produces trivial outputs can still perform fairly well in terms of typical paraphrasing  evaluation metrics, despite the output being of little use. Alternative metrics have been proposed to encourage more diverse outputs \cite{shima2011}. Typically evaluation of paraphrasing or text generation tasks is performed by using a similarity metric (usually some variant of BLEU \cite{bleu}) calculated against a held-out set \cite{Prakash,cnn2017,yu2017}.

\begin{figure}[t]
	\centering
	\begin{tabular}{l}\hline
		domain: Movies\\
		intent: MovieRating \\
		slots: \textbf{movie\_title} \\ \hline
		can children watch the movie \textbf{ movie\_title}\\
		can i watch the movie \textbf{movie\_title} with my son\\
		is \textbf{movie\_title} p. g. thirteen\\
		is \textbf{movie\_title} suitable for children\\\hline
		\\\hline
		domain: Movies\\
		intent: GetActorMovies \\
		slots: \textbf{genre}, \textbf{person\_name} \\ \hline
		give me \textbf{genre} movies starring \textbf{person\_name}\\
		suggest \textbf{genre} movies starring \textbf{person\_name}\\
		what \textbf{genre} movies is \textbf{person\_name} in\\
		what are \textbf{genre} movies with \textbf{person\_name}\\\hline
	\end{tabular}
	\caption{Example of template carrier phrases for two signatures $s$.}
	\label{fig:problem example}
\end{figure}

\section{Methodology}
\label{sec:methodology}

%This section describes our approach to the data augmentation problem, starting from the problem definition in the context of NLU, the proposed evaluation framework and the models we examined.

\subsection{Problem Definition}
\label{sec:problem_definition}

New capabilities for virtual agents are typically defined by a few phrases templates, also called \textit{carrier phrases}, as seen in Fig.~\ref{fig:problem example}. In carrier phrases the entity values, like the movie title `Batman', are replaced with their entity types, like \textbf{movie\_title}. These are also called \textit{slot values} and \textit{slot types}, respectively, in the NLU literature. For our generation task, these phrases define a category: all carrier phrases that share the same domain, intent and slot types are equivalent, in the sense that they prompt the same agent response. For the remainder of this paper we will refer to the combination of domain, intent and slot types as the \textit{signature} of a phrase.  Given a small amount of example carrier phrases for a given signature of a new capability (typically under 5 phrases), our goal is to generate additional semantically similar carrier phrases for the target signature. 

The core challenge lies in the very limited data we can work with. The low number of phrases per category is, as we will show, highly problematic when training some adversarial or reinforcement structures. Additionally the high number of categories makes getting an output of the desired signature harder, because many similar signatures will be very close in latent space.
%Therefore, the generated output text should be \textit{controlled} by the signature.

%In this example, all phrases would lead to a response by the agent informing the user of the age rating of the movie identified by movie\_title.

%While this may seem like an extremely low resource task, note that many signatures from the same domain share slot types and possibly similar intents and word content, therefore training a text generation model jointly across signatures should benefit generation for such low resource signatures.

 % vae with discriminator
\begin{figure}[h]
	\centering
	\includegraphics[width=0.49\textwidth]{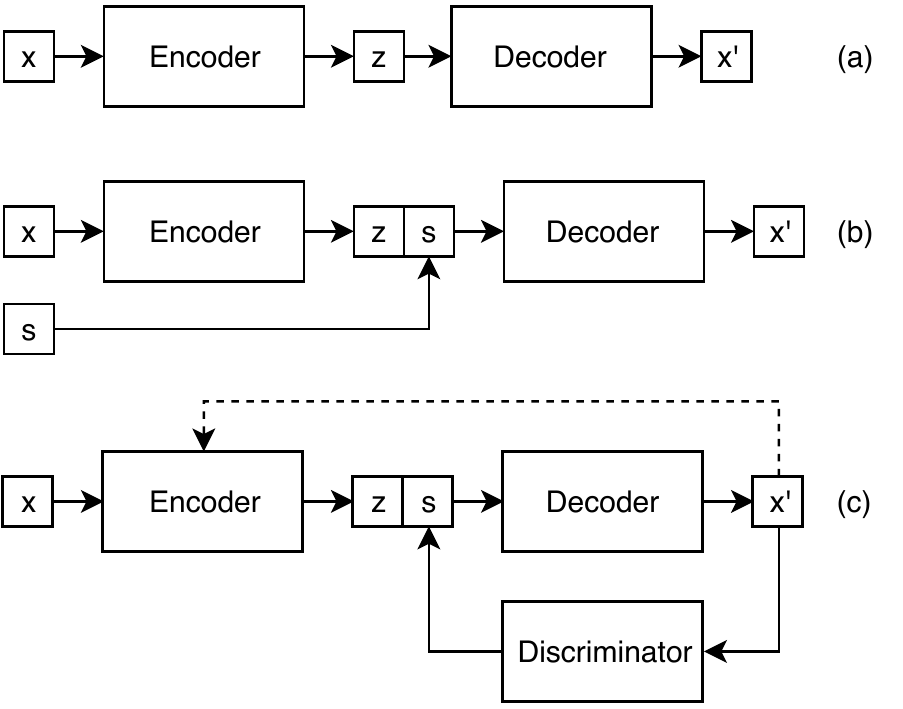}
	\caption{The variants of VAE we used: (a) VAE, (b) Conditional VAE (CVAE) and (c) VAE with discriminator}
	\label{fig:model_overview}
\end{figure}

\subsection{Generation models}
\label{sec:models}

Following is a short description of the models we evaluated for data generation. For all models we assume we have training carrier phrases $c_i \in D_{tr}^s$ across signatures $s$, and we pool together the data from all the signatures for training. The variational auto-encoders we used can be seen in Fig~\ref{fig:model_overview}.

\paragraph{Sequence to Sequence with Attention} 
Here, we use the seq2seq with global attention proposed in \cite{Luong2015} as our baseline generation model. The model is trained on all input-output pairs of carrier phrases belonging to the same signature $s$, e.g., $c_1, c_2 \in D_{tr}^s$. At generation, we aim to control the output by using an input carrier of the target signature $s$.

\paragraph{Variational Auto-Encoders} (VAEs) 
The VAE model can be trained with a paraphrasing objective, e.g., on pairs of carrier phrases $c_1, c_2 \in D_{tr}^s$, similarly to the seq2seq model. Alternatively, the VAE model can be trained with a reconstruction objective e.g., $c_1 \in D_{tr}$ can be both the input and the output. However, if we train with a reconstruction objective, during generation, we ignore the encoder and randomly sample the VAE prior $z$ (typically from a normal distribution). As a result, we have no control over the output signature distribution, and we may generate any of the signatures $s$ in our training data. This disadvantage motivates the investigation of two controlled VAE models.

\paragraph{VAE with discriminator} is a modification of a VAE proposed by \cite{Hu2017} for a similar task of controlled text generation. In this case, adversarial type of training is used by training a discriminator, i.e., a classifier for the category (signature \textbf{s}), to explicitly enforce control over the generated output. The network is trained in steps, with the VAE trained first, then the discriminator is attached and the entire network re-trained using a sleep-wake process. We tried two variations of this, one training a VAE, another training a CVAE, before adding the discriminator. Note that control over the output depends on the discriminator performance. While this model worked well for controlling between a small number of output categories as in \cite{Hu2017}, our setup includes hundreds of signatures $s$, which posed challenges in achieving accurate control over the output phrases (Sec. \ref{sec:intrinsic_results}).

\paragraph{Conditional VAE (CVAE)} Inspired by \cite{Moyer2018} for invariant feature learning, we propose to use a CVAE based controlled model structure. Such structure is a modification on the VAE, where we append the desired category label, here signature \textbf{s}, in 1-hot encoding, to each step of the decoder without an additional discriminator as shown in ~\cite{Hu2017}. Note that the original conditional VAE has already been applied to controlled visual settings \cite{Yan2015}. It has been shown that by direct optimizing the loss, this model automatically learns a invariant representation \textbf{z} that is independent of the category (signature $s$ \cite{Moyer2018}) although no explicit constraint is forced. We propose to use this model in our task, because it is easy to train (no wake-sleep or adversarial training), requires less data, and provides us a way to control the desired VAE output signature, by setting the desired signature encoding to \textbf{s}. Like the standard VAE, the CVAE can be trained either with a paraphrasing or with a reconstruction objective. If training with reconstruction, during generation we randomly sample from \textbf{z} but can control the output signature by setting \textbf{s}.

All model encoders and decoders are GRUs. For the discriminator we tried CNN and LSTM with no significant performance differences.

\begin{table*}[h]
	\centering
	\renewcommand{\tabcolsep}{5pt}
	%    \begin{tabular}{c|l|rrrrrlc}
	%    domain  & subset & carriers & signatures & slots & words & pairs \\\hline
	%    \multirow{3}[0]{*}{Movies} & train & 1,382  & 179   & 21    & 353   & 23,252 \\
	%          & dev   & 622   & 109   & 15    & 292   & 4,696 \\
	%          & test  & 520   & 69    & 10    & 254   & 4,778 \\\hline
	%    \multirow{3}[0]{*}{All} & train & 5,651  & 767   & 141   & 685   & 106,698 \\
	%          & dev   & 1,858  & 353   & 92    & 486   & 23,514 \\
	%          & test  & 1,855  & 340   & 84    & 471   & 22,944 \\
	%    \end{tabular}%

	\begin{tabular}{c|l|r|r|r|r}
		domain  & subset & carriers & signatures & slots & words \\\hline
		\multirow{3}[0]{*}{Movies} & train & 1,382  & 179   & 21    & 353   \\
		& dev   & 622   & 109   & 15    & 292    \\
		& test  & 520   & 69    & 10    & 254   \\\hline
		\multirow{3}[0]{*}{Live Entertainment}  & train & 4269  & 588   & 120    & 332   \\
		& dev   & 1236   & 244   & 77    & 194    \\
		& test  & 1335   & 271    & 74    & 217   \\\hline
		\multirow{3}[0]{*}{All} & train & 5,651  & 767   & 141   & 685   \\
		& dev   & 1,858  & 353   & 92    & 486   \\
		& test  & 1,855  & 340   & 84    & 471   \\
	\end{tabular}%    
	
	\caption{Data distribution and splits. `All' contains the combined Movie and Entertainment live datasets}
	\label{tab:data}%
\end{table*}%

\begin{table}[h]
	\centering
	\begin{tabular}{c|l|r|r}
		domain & subset & carriers & intents  \\\hline
		\multirow{4}[0]{*}{All} & train & 5651  & 136   \\
		& dev   & 1858  & 101    \\
		& train+dev & 7509  & 136    \\
		& test  & 1855  & 94     \\\hline
		Movies & test  & 520   & 37  \\
	\end{tabular}%
	\label{tab:extrinsicdata}%
	\caption{Data distribution and splits for the extrinsic task.}
\end{table}%

\section{Datasets}
\label{sec:data}

We experiment on two datasets collected for Alexa, a commercial artificial agent.

\paragraph{Movie dataset} It contains carrier phrases that are created as part of developing new movie-related functionality. It is composed of 179 signatures defined with an average of eight carrier phrases each. This data represents a typical new capability that starts out with few template carriers phrases, and we use it to examine if this low resource dataset can benefit from synthetic data generation.

\paragraph{Live entertainment dataset} It contains live customer data from  deployed entertainment related capabilities (music, books, etc), selected for their semantic relevance to movies. These utterances were de-lexicalized by replacing slot values with their respective slot types. We used a frequency threshold to filter out rare carrier phrases, and ensure a minimum number of three carrier phrases per signature. %This data represents typical live usage across functionalities and was used, along with movie phrases, to train and evaluate the generation models.

Table~\ref{tab:data} shows the data splits for the movie, live entertainment and `all' datasets, the latter containing both movies and live entertainment data, including the number of signatures, slot types and unique non-slot words in each set. While the data splits were stratified, signatures with fewer than four carriers were placed only in the train set, leading to the discrepancy in signature numbers across partitions.

\section{Experiments}
\label{sec:experiments}

\subsection{Experimental setup}
\label{sec:setup}

At the core of our data augmentation task lies the question ``what defines a good training data set?''. We can evaluate aspects of the generated data via synthetic metrics, but the most reliable method is to generate data for an extrinsic task and evaluate any improvements in performance. In this paper we employ both methods are reporting results for intrinsic and extrinsic evaluation metrics.

For the intrinsic evaluation, we train the data generator either only on movie data or on `all' data (movies and entertainment combined), using the respective dev sets for hyper-parameter tuning. During generation, we similarly consider either the movies test set, or the `all' test set, and aim to generate \textit{ten synthetic phrases} per test set phrase. VAE type generators can be trained for paraphrasing ($c1\rightarrow c2$) or reconstruction ($c1\rightarrow c1$). During generation, sampling can be performed either from the prior, e.g., by ignoring the encoder and sampling $z\sim\mathcal{N}(0,I)$ to generate an output, or from the posterior e.g., using $c_1$ as input to the encoder and producing the output $c2$. Note that not all combinations are applicable to all models. Those applicable are shown in Table~\ref{tab:maxresults_test}, where `para', `recon', `prior' and `post' denote paraphrasing, reconstruction, prior and posterior respectively. Special handling was required for a VAE with reconstruction training and prior sampling, where we have no control over the output signature. To solve this, we compared each output phrase to every signature in the train set (via BLEU4 \cite{bleu}) and assigned it to the highest scoring signature. Some sample output phrases can be seen in Fig.~\ref{fig:sample_outputs}.

\iffalse
As an example, assume a signature $s$ for which we have 9 training and 3 test samples and we want to generate data using a CVAE and sampling from the prior distribution.
Sampling from the prior means we will ignore the input test phrases instead giving the model the desired signature $s$ and 30 randomly generated vectors $z$ (10 per test phrase) sampled from a normal distribution $\mathcal{N}(0,I)$, producing 30 artificial phrases.
For these phrases we will calculate three corpus-level BLEU4 scores, against the train set, test set and the generated set itself, by using all phrases with the same signature as multiple references (minus the input phrase or currently evaluated generated phrase).
These BLEU4 scores will be used to calculate the corresponding intrinsic accuracy, novelty and diversity metrics.
Note that, because we are using multiple references, the BLEU-derived metrics do not require the generated phrases to perfectly match any references.
The number of references (phrases per signature) is fairly low and variable, but comparable to the number of references in Machine Translation and paraphrasing tasks.
In the experiments shown, where we have multiple signatures, all outputs and references are collected and BLEU4 scores are calculated over the entire corpus (not per signature), even though only phrases with the same \textit{assumed} signature are ever compared.
\fi

\begin{figure*}[t]
	\centering
	\begin{tabular}{ll}\hline
		& Model: S2S\\
		& Input: i \textbf{ negation } feel like watching a movie with \textbf{person\_name} \\\hline
		\parbox[t]{5mm}{\multirow{7}{*}{\rotatebox[origin=c]{90}{outputs}}} 
		& i \textbf{negation} like movies by \textbf{person\_name}\\
		& i \textbf{negation} feel like watching movies by \textbf{person\_name}\\
		& i \textbf{negation} feel like watching a movie by \textbf{person\_name}\\
		& i \textbf{negation} like \textbf{person\_name}\\
		& i \textbf{negation} feel like watching a movie\\
		& i \textbf{negation} want to talk about \textbf{person\_name}\\
		& no i \textbf{negation} like \textbf{person\_name} movies \\\hline
		\\\hline
		& Model: VAE, sampling from prior distribution\\
		& Input: GetMovieAwards (intent) - \textbf{award\_title}, \textbf{movie\_title} (slots)\\ \hline
		\parbox[t]{5mm}{\multirow{7}{*}{\rotatebox[origin=c]{90}{outputs}}}
		& did \textbf{movie\_title} win an \textbf{award\_title}\\
		& any \textbf{award\_title} won by \textbf{movie\_title}\\
		& tell me any \textbf{award\_title} which \textbf{movie\_title} won\\
		& was \textbf{movie\_title} nominated for an \textbf{award\_title} the movie \textbf{movie\_title}\\
		& any \textbf{award\_title} for \textbf{movie\_title}\\
		& what are the \textbf{award\_title} which \textbf{movie\_title} won\\
		& give me any \textbf{award\_title} the movie was nominated for\\ \hline
		\\\hline
		& Model: CVAE, sampling from prior distribution\\
		& Input: GetActorMovies (intent) - \textbf{genre}, \textbf{person\_name} (slots)\\ \hline
		\parbox[t]{5mm}{\multirow{7}{*}{\rotatebox[origin=c]{90}{outputs}}}
		& give me \textbf{genre} movies starring \textbf{person\_name}\\
		& show me other \textbf{genre} movies with \textbf{person\_name} in it\\
		& what are the \textbf{genre} movies that \textbf{person\_name} starred in\\
		& tell me \textbf{genre} movies starring \textbf{person\_name}\\
		& what are \textbf{genre} movies with \textbf{person\_name}\\
		& \textbf{genre} movies starring \textbf{person\_name}\\
		& suggest \textbf{genre} movies starring \textbf{person\_name}\\
		\hline
		
	\end{tabular}
	\caption{Sample output phrases. The S2S model and all posterior sampling models use a phrase as an input. For prior sampling the desired signature is the model input.}
	\label{fig:sample_outputs}
\end{figure*}

%Note that not all combinations are applicable to all models, e.g., it does not make sense to sample prior in seq2seq models neither to sample from prior a VAE model trained with paraphrasing loss. 

%we consider three setups when training a data generator: (a) train on only movie data and generate for the movie domain, (b) train on all data and generate for the movie domain, (c) train on all data and generate for all domains.

% Table generated by Excel2LaTeX from sheet 'Summary-Test'
\begin{table*}[th]
  \centering
%    \begin{small}
    \renewcommand{\tabcolsep}{5pt}
    \begin{tabular}{c|l|r|r|r|r|r|r|r|r}
    \multicolumn{2}{c}{} & \multicolumn{1}{|c}{S2S} & \multicolumn{2}{|c}{VAE} & \multicolumn{1}{|c}{VAE+DISC} & \multicolumn{3}{|c}{CVAE} & \multicolumn{1}{|c}{CVAE+DISC} \\ \hline
    \multicolumn{2}{c|}{training} & para  & recon & para  & recon & recon & recon & para  & recon \\ \hline
    \multicolumn{2}{c|}{sampling} & post  & prior & post  & prior & prior & post  & post  & prior \\ \hline
    \multirow{2}[0]{*}{accuracy} & BLEU4 & 0.86  & 0.91  & 0.24  & 0.42  & 0.91  & 0.88  & 0.90  & 0.11 \\
%          & match rate & 0.06  & 0.09  & 0.00  & 0.00  & 0.04  & 0.04  & 0.03  & 0.00 \\
          & slot c.o. & 0.84  & 0.95  & 0.02  & 0.12  & 0.98  & 0.93  & 0.95  & 0.01 \\\hline
%    \multirow{2}[0]{*}{diversity} & 1-BLEU4 & 0.12  & 0.37  & 1.66  & 1.67  & 0.27  & 0.45  & 0.66  & 0.37 \\
    
    \multirow{2}[0]{*}{diversity} & 1-BLEU4 & 0.06 & 0.19	& 0.83 & 0.84 & 0.14 & 0.23 & 0.33 & 0.19 \\

          & uniq. rate & 0.58  & 0.68  & 0.98  & 0.76  & 0.44  & 0.56  & 0.68  & 0.97 \\\hline
%    fluency & -ppl  & -31.50 & -43.54 & -46.69 & -21.32 & -42.18 & -41.76 & -41.31 & -21.83 \\\hline
    \multirow{2}[0]{*}{novelty} & 1-BLEU4 & 0.25  & 0.07  & 0.75  & 0.98  & 0.04  & 0.12  & 0.21  & 0.99 \\
          & 1-match rate & 0.89  & 0.76  & 0.99  & 1.00  & 0.32  & 0.50  & 0.59  & 1.00 \\\hline
    \end{tabular}%
%    \end{small}
  \caption{Best performance per metric for each model when applied to `all' domains.}
  
  \label{tab:maxresults_test}%
\end{table*}%

To examine the usefulness of the generated data for an extrinsic ask, we perform intent classification, a standard task in NLU.
%\colorbox{yellow}{[ref]}. 
Our classifier is a BiLSTM model. We use the same data as for the data generation experiments (see Table~\ref{tab:data}), and group our class labels into intents (as opposed to signatures), which leads to classifying 136 intents in the combined movies and entertainment data (`all'). Our setup follows two steps: First, the data generators are trained on `all' train sets, and used to generate phrases for the dev sets (`all' and movies). Second, the intent classifier is trained on the `all' train and dev sets (baseline), vs the combination of `all' train, dev and generated synthetic data, which is our proposed approach. We evaluate on the `all' and movies test sets, and use macro-averaged F-score across all intents as our metric.

%To validate the usefulness of the generated data to external tasks we use it as extra data for a classification task. We use a biLSTM model to classify carrier phrases to intents, using the same data as the used for the generation experiments, and compare how performance is affected by adding the synthetic data to the real data.
%The real data are the same as in Table~\ref{tab:data} but this time we are not using signatures, just intents (slots ignored) which leads to fewer categories, 136 for all data, and a more manageable task.

%The data generators are trained on all domains train or train \& dev sets, the former scenario applying to paraphrasing models and the latter to reconstruction models. Then these models are used to generate phrases for the all domain test set.
%The extrinsic classifier is trained on the all domain train \& dev sets plus one set of synthetic phrases, then used to classify the all domain or only the movie test set.
%The evaluation metric used is macro-averaged f-score across all intents.

\subsection{Intrinsic evaluation}
\label{sec:intrinsic_results}

%\iffalse
\begin{figure}[h]
	\centering
	\includegraphics[width=0.35\textwidth]{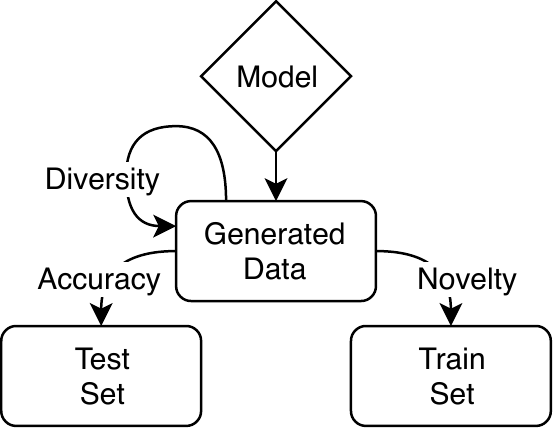}
	\caption{Intrinsic evaluation overview}
	\label{fig:eval_overview}
\end{figure}
%\fi

To evaluate the generated data we use an ensemble of evaluation metrics attempting to quantify three important aspects of the data: (1) how accurate or relevant the data is to the task, (2) how diverse the set of generated phrases is and (3) how novel these synthetic phrases are.
Intuitively, a NLG system can be very accurate - generate valid phrases of the correct signature - while only generating phrases from the train set or while generating the same phrase multiple times for the same signature; either of these scenaria would not lead to useful data.
To evaluate accuracy we compare the generated data to a held out \textit{test set} using BLEU4 \cite{bleu} and the slot carry-over rate, the probability that a generated phrase contains the exact same slot types as the target signature $s$.
To evaluate novelty we compare the generated data to the train set of the generator, using 1-BLEU4 (where higher is better) and 1-Match rate, where the match rate is the chance that a perfect match to a generated phrase exists in the train set. These scores tell us how different, at the lexical level, the generated phrases are to the phrases that already exist in the train set.
Finally, to evaluate diversity we compare the phrases in the generated data to each other, using again 1-BLEU4 and the unique rate, the number of unique phrase produced over the total number of phrases produced. These scores indicate how lexically different the generated phrases are to each other.
Figure~\ref{fig:eval_overview} shows the set comparisons made to generate the intrinsic evaluation metrics.
Note that these metrics mostly evaluate surface forms; we expect phrases generated for the same signature to be \textit{semantically} similar to phrases with the same signature in the train set and to each other, however we would like them to be \textit{lexically} novel and diverse.

Table \ref{tab:maxresults_test} presents the intrinsic evaluation results, where generators are trained and tested on `all' data, for the best performing model per case, tuned on the dev set. First, note the slot carry over (slot c.o.), which can be used as a sanity check measuring the chance of getting a phrase with the desired slot types. Most models reach 0.8 or higher slot c.o. as expected, but some fall short, indicating failure to produce the desired signature. The failure for VAE and CVAE models with discriminators is most notable, and can be explained by the fact that we have a large number of train signatures ($\sim$800) and too few samples per signature (mean 8, median 4), to accurately train the discriminator. We verified that the discriminator overall accuracy does not exceed 0.35. The poor discriminator performance leads to the decoder not learning how to use signature \textbf{s}. The failure of VAE with posterior sampling is similarly explained by the large number of signatures: the signatures are so tightly packed in the latent space, that the variance of sampling \textbf{z} is likely to result in phrases from similar but different signatures. 

\begin{figure}[h]
	\centering
	\includegraphics[width=0.49\textwidth]{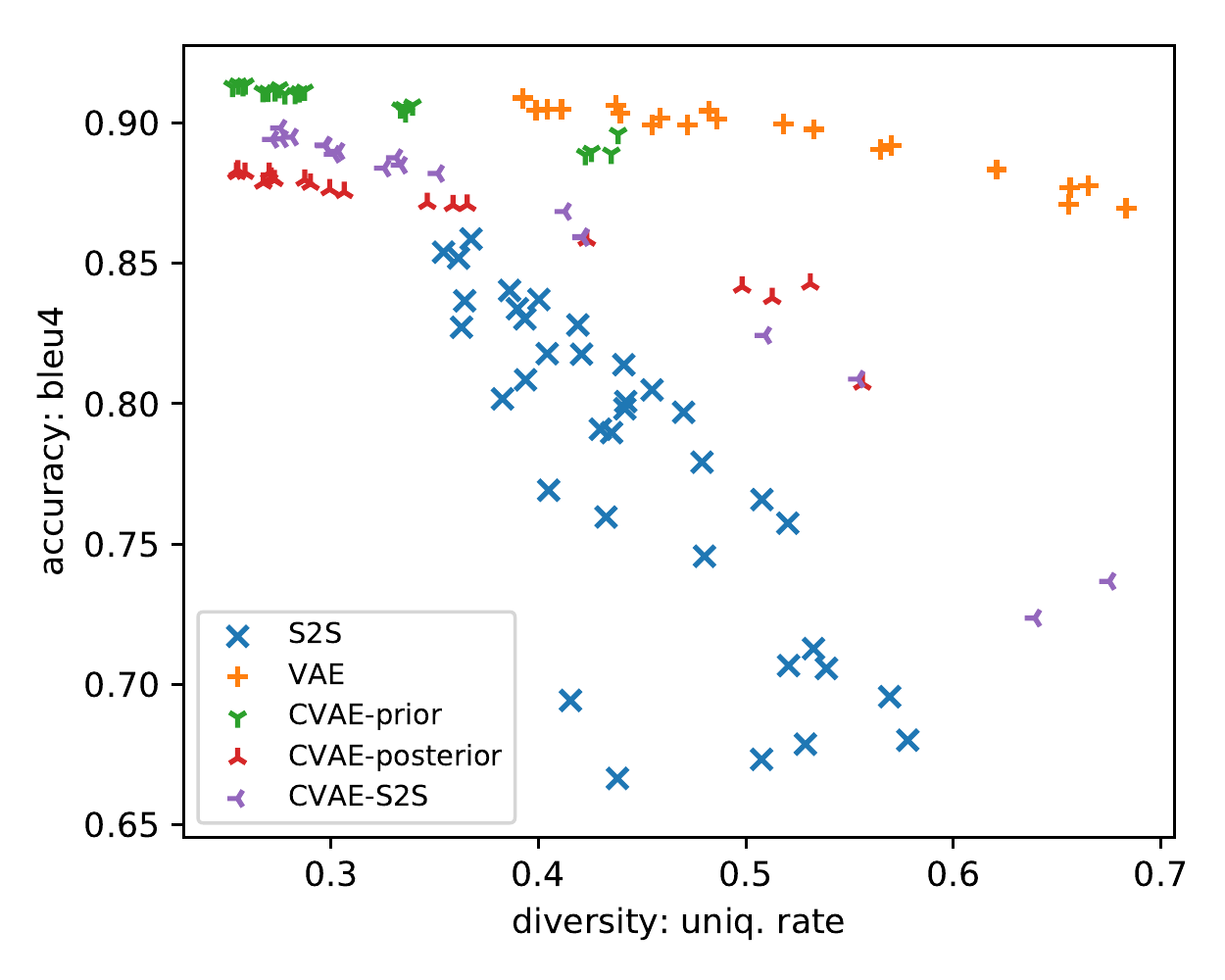}
	\caption{Intrinsic accuracy BLEU4 as a function of intrinsic diversity for the `all' test set for multiple hyper-parameter combinations of each model.}
	\label{fig:res1}
\end{figure}

This sanity check leaves us with five reasonably performing models: S2S, VAE trained for reconstruction and sampled from the prior and CVAE with multiple training and sampling strategies. Overall, these models achieve high accuracy with respect to the slot c.o. and BLEU4 metrics, assisted by the rather limited vocabulary of the data. To examine the trade-offs between the models, in Fig.~\ref{fig:res1}, we show the accuracy BLEU4 as a function of diversity unique rate, i.e., how many different phrases we generated. Each point is a model trained with different hyper-parameter settings, across relevant hyper-parameters, network component dimensionalities etc. As expected, diversity is negatively correlated with accuracy. We make similar observations for novelty metrics (plots omitted for brevity), i.e., diversity and novelty are negatively correlated to accuracy within the hyper-parameter constraints of each model. However the trade-off is not equally steep for all models. Across our experiments the VAE and CVAE models with reconstruction training and prior sampling provided the most favorable trade-offs with CVAE being the best option for very high accuracy, as seen in Fig.~\ref{fig:res1}.

%The first thing to note is slot carry over, the chance of getting a phrase with the desired slot types, since it is an easy sanity check. Most models can reach 0.8 or higher slot carry-over as expected, but some fall very short, indicating that they fail to produce phrases of the desired signature. The VAE and CVAE models with discriminators are most notable, but easy to explain: we have too few samples per category (mean 8, median 4) to effectively train the discriminator. We conducted a separate round of experiments investigating the performance of the discriminator and found we could not achieve accuracy higher than 0.35 for all domains. Apparently the poor performance of the discriminator leads to the decoder learning very little of how to use \textbf{c}. This is easy to attribute to the discriminator performance, since the CVAE models (that are given perfect \textbf{c} for training) perform very well. For a similar reason, the very large number of categories, sampling from the posterior of a VAE fails: the categories are tightly packed in the latent space, to the point that the variance of sampling \textbf{z} is likely to result in samples from similar but different signatures. Again this is easy to attribute to the category density, since the S2S model (where sampling from the latent space has no variance) performs well.

In Table~\ref{tab:moviechange}, we show intrinsic results on the movies test set. For brevity, we show the mean relative change for the best performing models for each metric, computed between using only movie data to train the generators vs using the combined `all' data. In the latter case, the live entertainment data is added to train a more robust generator for movies. As expected, we notice a small loss in accuracy (-1.9 \% rel. change on average for BLEU4) when using the `all' data for generator training, but also a significant gain in diversity and novelty of the movie generated data (121 \% and 153 \% rel. change on average respectively for 1-BLEU4). Overall, the reconstruction VAE and CVAE models achieve the best results and have favorable performance trade-offs when using `all' data to enrich movie data generation.

%Apart from all domains we also experimented on using just the movie domain data for training and comparing movie data vs all data when applied only to generate phrases for the movie domain.Table~\ref{tab:moviechange} shows the mean change in performance across all models when adding the data from all other domains to the movie data. The results are predictable in that we lose accuracy but gain diversity and novelty, but the trade-off is more favorable than expected: we lose very little accuracy for large gains in diversity and novelty, a very positive outcome.

%Overall the intrinsic metrics show high performance, with the reconstruction VAE and CVAE taking top honors and what appear to be favorable performance trade-offs when using data from other domains.

\begin{table*}[t]
	\centering
%	\begin{small}
		\renewcommand{\tabcolsep}{5pt}
		\begin{tabular}{l|c|c|c|c}
			metric     & \multicolumn{4}{c}{ \% change} \\ \hline
			& CVAE  & CVAE & CVAE & VAE  \\ 
			& prior & posterior & s2s & prior \\\hline
			\multicolumn{5}{c}{accuracy} \\ \hline
			BLEU4 & -0.7\%	&	-0.7\%	&	-4.3\%	&	-2.0\% \\
%			match rate & 74.7\%	&	96.2\%	&	97.8\%	&	26.7\% \\
			slot c.o. & -2.7\%	&	-3.1\%	&	-9.5\%	&	-5.9\% \\ \hline
			\multicolumn{5}{c}{diversity} \\ \hline
			1-BLEU4 &  112.1\%	&	103.6\%	&	133.7\%	&	134.6\% \\
			uniq. rate & 19.6\%	&	19.7\%	&	32.7\%	&	37.1\% \\ \hline
			\multicolumn{5}{c}{novelty} \\ \hline
			1-BLEU4 & 147.1\%	&	40.7\%	&	222.1\%	&	201.2\% \\
			1-match rate &  113.5\%	&	36.7\%	&	145.4\%	&	90.1\% \\\hline
		\end{tabular}%
%	\end{small}
	\caption{Relative change when adding more training data for generator training (movies only vs `all') across evaluation metrics on the movies test set }
	\label{tab:moviechange}%
\end{table*}%

\begin{figure*}[ht]
	\centering
	\renewcommand{\tabcolsep}{1pt}
	\begin{tabular}{cc}
		%    \includegraphics[width=0.41\textwidth]{extrinsic0-acc-all_delex.pdf} & 
		%    \includegraphics[width=0.41\textwidth]{extrinsic0-acc-moviebotgrammar_delex.pdf}\\
		%    (a) &
		\includegraphics[width=0.50\textwidth,height=0.26\textheight]{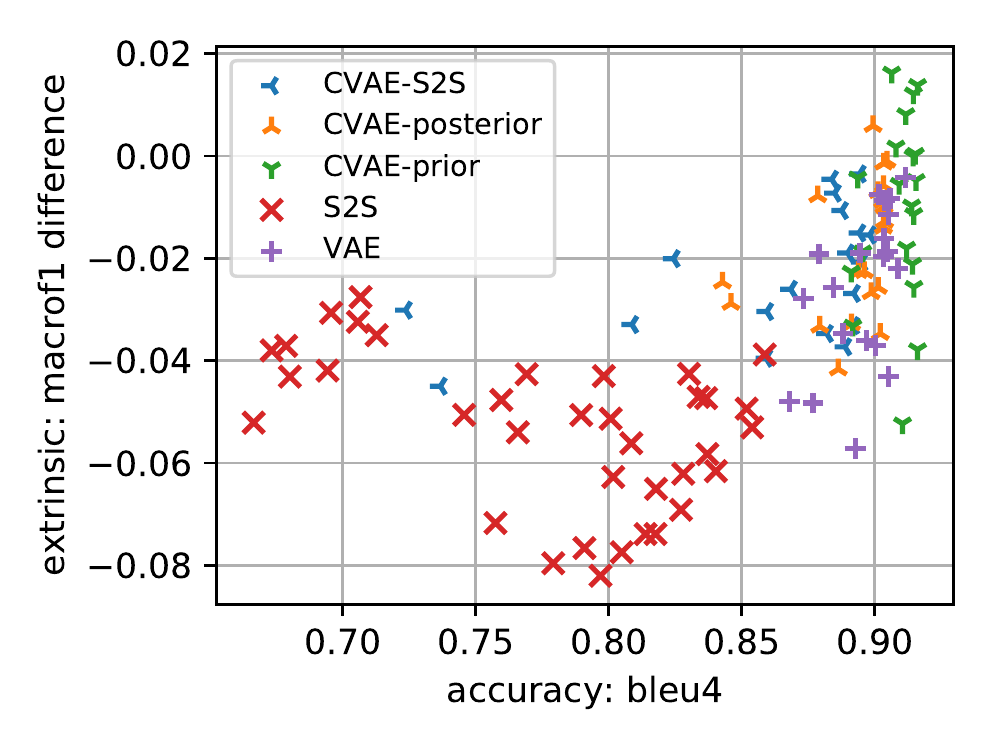} &
		\includegraphics[width=0.50\textwidth,height=0.26\textheight]{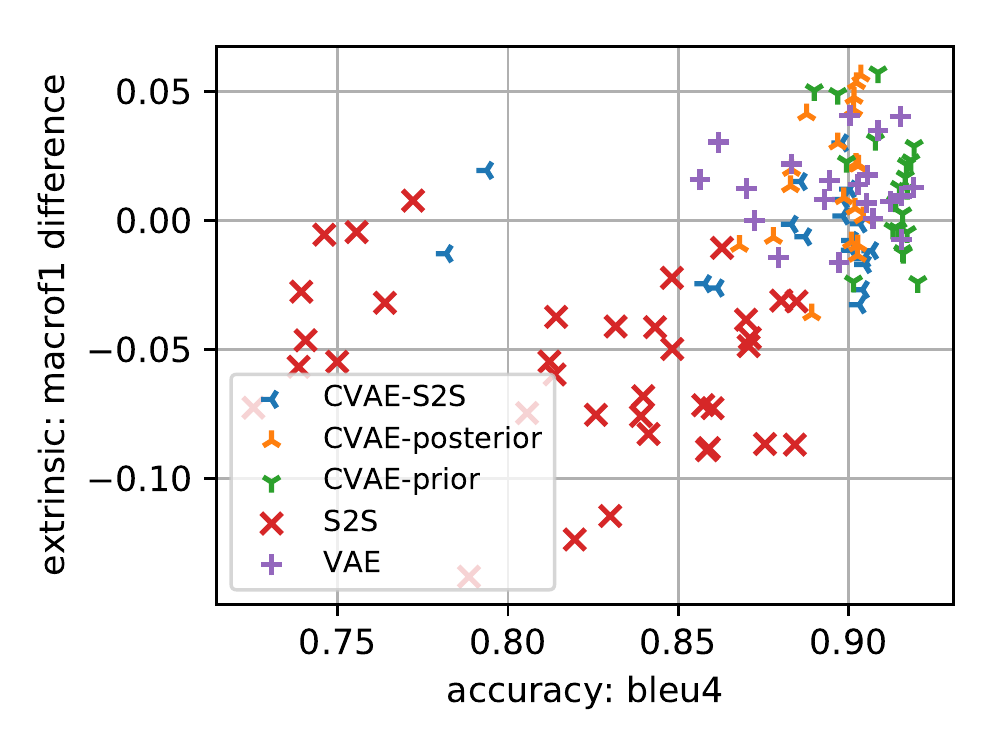} \\ 
		(a) & (b)\\
	\end{tabular}
	\caption{Extrinsic task performance change as a function of intrinsic accuracy BLEU4 for (a) all domains and (b) the movies domain. The $y$ axis represents how much performance improved or deteriorated after adding synthetic data to the train set.}
	%    \caption{Extrinsic task performance change as a function of intrinsic BLEU4 for the (a) all domains and (b) movies experiments.}
	\label{fig:extrinsic1}
\end{figure*}

\subsection{Extrinsic Evaluation}
\label{sec:extrinsic_results}

%containing combined entertainment and movies domains, while the lower plot presents results on the movies domain. 

In Figure~\ref{fig:extrinsic1} we present the change in the F1 score for intent classification when adding the generated data into the classifier training (compared to the baseline classifier with no generated data) as a function of the intrinsic BLEU4 accuracy metric. The plot presents results on the movies test set. Each point is a model trained with different hyper-parameters and the line $y=0$  represents zero change from baseline, while models over this line represent improvement. Some hyper-parameter choices clearly lead to sub-optimal results, but they are included to show the relationship between intrinsic and extrinsic performance across a wider range of conditions. We notice that many generators produce useful synthetic data that lead to improvement in intent classification, with the best performing ones being the CVAE models with around 5\% absolute improvement in F-score on the movie test set ($p < 0.01$). This is an encouraging results, as it verifies the usefulness of the generated data for improving the extrinsic low resource task. 
%It also validates the proposed intrinsic BLEU4 accuracy metric which is correlated with the extrinsic F-1 improvement. 
For the `all' test set experiments, the improvement is less pronounced, with maximum gain from synthetic data being around 2\%, again for the  CVAE models. This smaller improvement could be because this test set is not as low resource (roughly twice as many train carriers phrases per intent on average, 41.55 instead of 24.25), therefore harder to improve using synthetic data. Note that the baseline F1 scores (no synthetic data) are 0.58 for movies and 0.60 for the `all' test set.

\iffalse
\begin{figure}[h]
	\centering
	\renewcommand{\tabcolsep}{1pt}
	\begin{tabular}{c}
		%    \includegraphics[width=0.41\textwidth]{extrinsic0-acc-all_delex.pdf} & 
		%    \includegraphics[width=0.41\textwidth]{extrinsic0-acc-moviebotgrammar_delex.pdf}\\
%    (a) &
     \raisebox{-0.5\height}{\includegraphics[width=0.49\textwidth]{extrinsic0-macrof1-all_delex.pdf}} \\ 
     (a) \\
%	(b)	&
	 \raisebox{-0.5\height}{\includegraphics[width=0.49\textwidth]{extrinsic0-macrof1-moviebotgrammar_delex.pdf}} \\
	 (b)\\
	\end{tabular}
	\caption{Extrinsic task performance change as a function of intrinsic accuracy BLEU4 for (a) all domains and (b) the movies domain. The $y$ axis represents how much performance improved or deteriorated after adding synthetic data to the train set.}
	%    \caption{Extrinsic task performance change as a function of intrinsic BLEU4 for the (a) all domains and (b) movies experiments.}
	\label{fig:extrinsic1}
\end{figure}
\fi

We investigate the correlation between the intrinsic metrics and the extrinsic F score by performing Ordinary Least Squares (OLS) regression between the two types of metrics, computed on the `all' test set. We find that intrinsic accuracy metrics like BLEU4 and slot c.o. have significant positive correlation with macro F ($R^2$ of 0.31 and 0.40 respectively, $p \approx 0$) across all experiments/models, though perhaps not as high as one might expect.  We also computed via OLS the combined predictive power of all intrinsic metrics for predicting extrinsic F, and estimated an $R^2$ coefficient of 0.53 ($p \approx 0$). The diversity and novelty metrics add a lot of predictive power to the OLS model when combined with accuracy metrics, raising R$^2$ from 0.40 to 0.53, validating the need to take these aspects of NLG performance into account. However, intrinsic diversity and novelty are only good predictors of extrinsic performance when combined with accuracy, so they only become significant when comparing models of similar intrinsic accuracy.

\section{Conclusions}
\label{sec:conclusions}

We described a framework for controlled text generation for enriching training data for new NLU functionality. Our challenging text generation setup required control of the output phrases over a large number of low resource signatures of NLU functionality. We used intrinsic metrics to evaluate the quality of the generated synthetic data in terms of accuracy, diversity and novelty. We empirically investigated variational encoder-decoder type models and proposed to use a CVAE based model, which yielded the best results, being able to generate phrases with favorable accuracy, diversity and novelty trade-offs. We also demonstrated the usefulness of our proposed methods by showing that the synthetic data can improve the accuracy of an extrinsic low resource classification task.

% In the future we will be investigating the use of unlabeled data as part of the training process, as well as different, more guided, sampling strategies.

\section{Acknowledgments} 
This work was performed while Nikolaos Malandrakis was at Amazon Alexa AI, Sunnyvale.

\bibliography{mybib}
\bibliographystyle{acl_natbib}

\end{document}